\documentclass[sigconf]{acmart}

\AtBeginDocument{%
  \providecommand\BibTeX{{%
    \normalfont B\kern-0.5em{\scshape i\kern-0.25em b}\kern-0.8em\TeX}}}

\setcopyright{acmcopyright}
\copyrightyear{2018}
\acmYear{2018}
\acmDOI{10.1145/1122445.1122456}

\copyrightyear{2020} 
\acmYear{2020} 
\setcopyright{acmcopyright}\acmConference[MM '20]{Proceedings of the 28th ACM International Conference on Multimedia}{October 12--16, 2020}{Seattle, WA, USA}
\acmBooktitle{Proceedings of the 28th ACM International Conference on Multimedia (MM '20), October 12--16, 2020, Seattle, WA, USA}
\acmPrice{15.00}
\acmDOI{10.1145/3394171.3413896}
\acmISBN{978-1-4503-7988-5/20/10}

\usepackage{epsfig}
\usepackage{graphicx}
\usepackage{amsmath}
\usepackage{amssymb}

\usepackage{algorithm, algpseudocode}
\usepackage{multirow}
\usepackage{caption}
\usepackage{comment}
\usepackage{amsmath}
\usepackage{amssymb}
\definecolor{citecolor}{RGB}{119,185,0} 

\usepackage{pifont}
\usepackage{color}

\newlength\savewidth\newcommand\shline{\noalign{\global\savewidth\arrayrulewidth
  \global\arrayrulewidth 1pt}\hline\noalign{\global\arrayrulewidth\savewidth}}

\def\eg{\emph{e.g.}} 
\def\ie{\emph{i.e.}} 
\def\etal{\emph{et~al.}} 

\begin{document}
\fancyhead{}
\title{ University-1652: A Multi-view Multi-source Benchmark \\ for Drone-based Geo-localization}

\author{  
Zhedong Zheng \quad Yunchao Wei \quad Yi Yang$^{*}$  } 
\affiliation{%
  \institution{SUSTech-UTS Joint Centre of CIS,
Southern University of Science and Technology \\ ReLER, AAII, University of Technology Sydney}
  }
\email{zhedong.zheng@student.uts.edu.au, yunchao.wei@uts.edu.au, yi.yang@uts.edu.au}
\thanks{$^{*}$Corresponding author.}

\begin{abstract}
We consider the problem of cross-view geo-localization. 
The primary challenge is to learn the robust feature against large viewpoint changes. Existing benchmarks can help, but are limited in the number of viewpoints. Image pairs, containing two viewpoints, \eg, satellite and ground, are usually provided, which may compromise the feature learning. Besides phone cameras and satellites, in this paper, we argue that drones could serve as the third platform to deal with the geo-localization problem. In contrast to traditional ground-view images, drone-view images meet fewer obstacles, \eg, trees, and provide a comprehensive view when flying around the target place. To verify the effectiveness of the drone platform, we introduce a new multi-view multi-source benchmark for drone-based geo-localization, named University-1652. University-1652 contains data from three platforms, \ie, synthetic drones, satellites and ground cameras of $1,652$ university buildings around the world.
To our knowledge, University-1652 is the first drone-based geo-localization dataset and enables two new tasks, \ie, drone-view target localization and drone navigation.
As the name implies, drone-view target localization intends to predict the location of the target place via drone-view images. On the other hand, given a satellite-view query image, drone navigation is to drive the drone to the area of interest in the query.
We use this dataset to analyze a variety of off-the-shelf CNN features and propose a strong CNN baseline on this challenging dataset. The experiments show that University-1652 helps the model to learn viewpoint-invariant features and also has good generalization ability in real-world scenarios. 
\end{abstract}



\begin{teaserfigure}
    \centering
    \includegraphics[width=\linewidth]{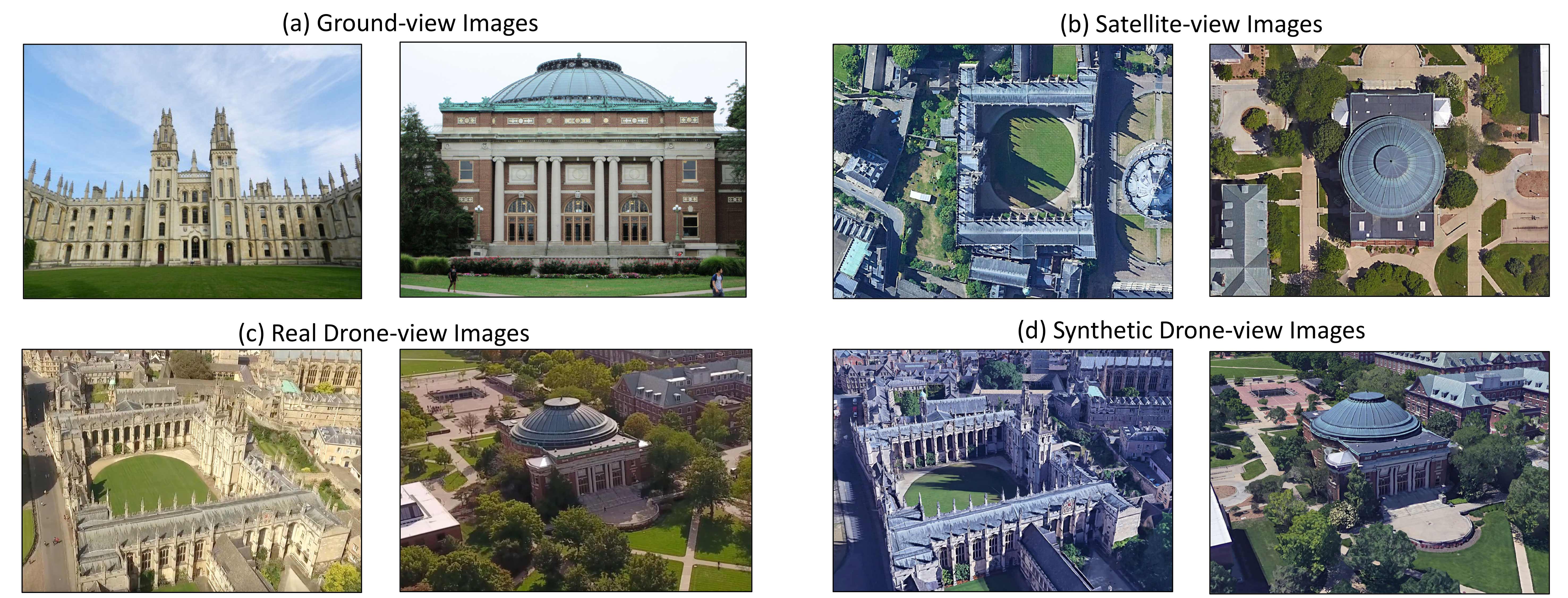}
    \vspace{-.3in}
    \captionof{figure}{
   It is challenging, even for a human, to associate (a) ground-view images with (b) satellite-view images. In this paper, we introduce a new dataset based on the third platform, \ie, drone, to provide real-life viewpoints and intend to bridge the visual gap against views. (c) Here we show two real drone-view images collected from public drone flights on Youtube \cite{Regal15,FlyLow}. (d) In practice, we use the synthetic drone-view camera to simulate the real drone flight. It is based on two concerns. First, the collection expense of real drone flight is unaffordable. Second, the synthetic camera has a unique advantage in the manipulative viewpoint. Specifically, the 3D engine in Google Earth is utilized to simulate different viewpoints in the real drone camera. 
    }
    \label{fig:0}
\end{teaserfigure}

\maketitle

\section{Introduction}
The opportunity for cross-view geo-localization is immense, which could enable subsequent tasks, such as, agriculture, aerial photography, navigation, event detection and accurate delivery \cite{zhu2018vision,brar2015drones,yu2019building}.
Most previous works regard the geo-localization problem as a sub-task of image retrieval \cite{philbin2007object,torii201524,arandjelovic2016netvlad,liu2019lending,tian2017cross,lin2015learning,yang2009ranking,wu2019progressive}. Given one query image taken at one view, the system aims at finding the most relevant images in another view among large-scale candidates (gallery). Since candidates in the gallery, especially aerial-view images, are annotated with the geographical tag, we can predict the localization of the target place according to the geo-tag of retrieval results. 

\begin{figure*}[t]
\begin{center}
    \includegraphics[width=1\linewidth]{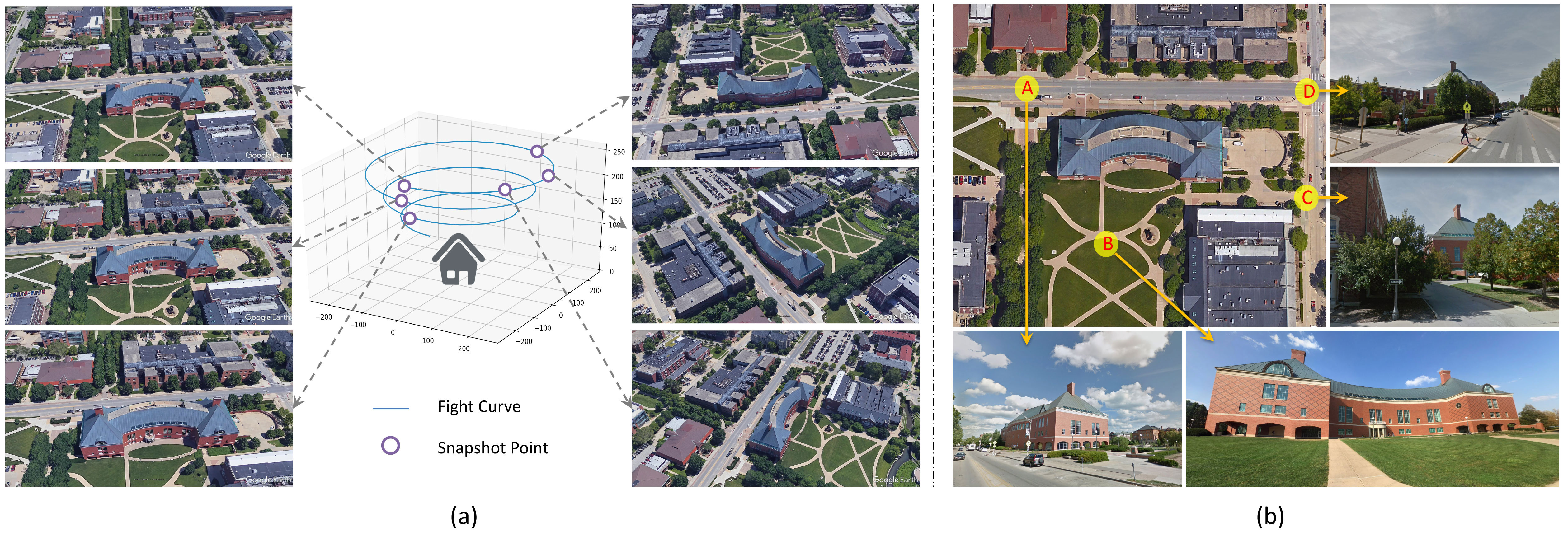}
\end{center}
\vspace{-.2in}
     \caption{(a) The drone flight curve toward the target building. When flying around the building, the synthetic drone-view camera could capture rich information of the target, including scale and viewpoint variants. (b) The ground-view images are collected from street-view cameras to obtain different facets of the building as well. It simulates real-world photos when people walk around the building.
     }\label{fig:spiral}
\end{figure*}

In general, the key to cross-view geo-localization is to learn a discriminative image representation, which is invariant to visual appearance changes caused by viewpoints. Currently, most existing datasets usually provide image pairs and focus on matching the images from two different platforms, \eg, phone cameras and satellites \cite{zhai2017predicting,liu2019lending}. 
As shown in Figure~\ref{fig:0} (a) and (b), the large visual difference between the two images, \ie, ground-view image and satellite-view image, is challenging to matching even for a human. The limited two viewpoints in the training set may also compromise the model to learn the viewpoint-invariant feature.

In light of the above discussions, it is of importance to (1) introduce a multi-view dataset to learn the viewpoint-invariant feature and bridge the visual appearance gap, and (2) design effective methods that fully exploit the rich information contained in multi-view data. With the recent development of the drone \cite{zhu2018vision,hsieh2017drone,li2017visual}, we reveal that the  drone could serve as a primary data collection platform for cross-view geo-localization (see Figure~\ref{fig:0} (c) and (d)). Intuitively, drone-view data is more favorable because drones could be motivated to capture rich information of the target place. When flying around the target place, the drone could provide comprehensive views with few obstacles. In contrast, the conventional ground-view images, including \emph{panorama}, inevitably may face occlusions, \eg, trees and surrounding buildings. 

However, large-scale real drone-view images are hard to collect due to the high cost and privacy concerns. In light of the recent practice using synthetic training data  \cite{Richter_2016_ECCV,liu2018pose,wu2018dcan,li2020metaparsing}, we propose a multi-view multi-source dataset called University-1652, containing synthetic drone-view images. University-1652 is featured in several aspects. First, it contains multi-view images for every target place. We manipulate the drone-view engine to simulate images of different viewpoints around the target, which results in $54$ drone-view images for every place in our dataset. 
Second, it contains data from multiple sources. Besides drone-view images, we also collect satellite-view images and ground-view images as reference. 
Third, it is large-scale, containing $50,218$ training images in total, and has $71.64$ images per class on average. The images in the benchmark are captured over $1,652$ buildings  of $72$ universities. More detailed descriptions will be given in Section~\ref{sec:dataset}.
Finally, University-1652 enables two new tasks, \ie, drone-view target localization and drone navigation. 

\noindent\textbf{Task 1: Drone-view target localization. (Drone $\rightarrow$ Satellite)} Given one drone-view image or video, the task aims to find the most similar satellite-view image to localize the target building in the satellite view. 

\noindent\textbf{Task 2: Drone navigation. (Satellite $\rightarrow$ Drone)}  Given one satellite-view image, the drone intends to find the most relevant place (drone-view images) that it has passed by. According to its flight history, the drone could be navigated back to the target place.

In the experiment, we regard the two tasks as cross-view image retrieval problems and compare the generic feature trained on extremely large datasets with the viewpoint-invariant feature learned on the proposed dataset. We also evaluate three basic models and three different loss terms, including contrastive loss \cite{lin2015learning,workman2015wide,zheng2016discriminatively}, triplet loss \cite{chechik2010large,deng2018triplet}, and instance loss \cite{zheng2017dual}. Apart from the extensive evaluation of the baseline method, we also test the learned model on real drone-view images to evaluate the scalability of the learned feature. Our results show that University-1652 helps the model to learn the viewpoint-invariant feature, and reaches a step closer to practice. Finally, the University-1652 dataset, as well as code for baseline benchmark, will be made publicly available for fair use.

\section{Related Work}

\subsection{Geo-localization Dataset Review}
Most previous geo-localization datasets are based on image pairs, and target matching the images from two different platforms, such as phone cameras and satellites. One of the earliest works \cite{lin2015learning} proposes to leverage the public sources to build image pairs for the ground-view and aerial-view images. It consists of 78k image pairs from two views, \ie, 45$^{\circ}$ bird view and ground view. 
Later, in a similar spirit, Tian \etal~ \cite{tian2017cross} collect image pairs for urban localization. Differently, they argue that the buildings could serve as an important role to urban localization problem, so they involve building detection into the whole localization pipeline. Besides, the two recent datasets, \ie, CVUSA \cite{zhai2017predicting} and CVACT \cite{liu2019lending}, study the problem of matching the panoramic ground-view image and satellite-view image. It could conduct user localization when Global Positioning System (GPS) is unavailable. The main difference between the former two datasets \cite{lin2015learning, tian2017cross} and the later two datasets \cite{zhai2017predicting,liu2019lending} is that the later two datasets focus on localizing the user, who takes the photo. In contrast, the former two datasets and our proposed dataset focus on localizing the target in the photo. Multiple views towards the target, therefore, are more favorable, which could drive the model to understand the structure of the target as well as help ease the matching difficulty. 
The existing datasets, however, usually provide the two views of the target place. Different from the existing datasets, the proposed dataset, University-1652, involves more views of the target to boost the viewpoint-invariant feature learning. 

\subsection{Deeply-learned Feature for Geo-localization}
Most previous works treat the geo-localization as an image retrieval problem. The key of the geo-localization is to learn the view-point invariant representation, which intends to bridge the gap between images of different views. With the development of the deeply-learned model, convolutional neural networks (CNNs) are widely applied to extract the visual features. 
One line of works focuses on metric learning and builds the shared space for the images collected from different platforms. Workman \etal~ show that the classification CNN pre-trained on the Place dataset \cite{zhou2017places} can be very discriminative by itself without explicitly fine-tuning \cite{workman2015location}. The contrastive loss, pulling the distance between positive pairs, could further improve the geo-localization results \cite{workman2015wide,lin2015learning}. Recently, Liu \etal~ propose Stochastic Attraction and Repulsion
Embedding (SARE) loss, minimizing the KL divergence between the learned and the actual distributions \cite{liu2019stochastic}.
Another line of works focuses on the spatial misalignment problem in the ground-to-aerial matching. Vo \etal~ evaluate different network structures and propose an orientation regression loss to train an orientation-aware network \cite{vo2016localizing}. Zhai \etal~ utilize the semantic segmentation map to help the semantic alignment \cite{zhai2017predicting}, and Hu \etal~ insert the NetVLAD layer \cite{arandjelovic2016netvlad} to extract discriminative features \cite{hu2018cvm}. Further, Liu \etal~ propose a Siamese Network to explicitly involve the spatial cues, \ie, orientation maps, into the training \cite{liu2019lending}. Similarly, Shi \etal~ propose a spatial-aware layer to further improve the localization performance \cite{shi2019optimal}. In this paper, since each location has a number of training data from different views, we could train a classification CNN as the basic model. When testing, we use the trained model to extract visual features for the query and gallery images. Then we conduct the feature matching for fast geo-localization.

\setlength{\tabcolsep}{6pt}
\begin{table*}
\small
\begin{center}
\begin{tabular}{l|c|c|c|c|c|c}
\hline
Datasets & University-1652 & CVUSA \cite{zhai2017predicting} & CVACT \cite{liu2019lending}  & Lin \etal \cite{lin2015learning} & Tian \etal \cite{tian2017cross} & Vo \etal \cite{vo2016localizing}\\
\shline
\#training & 701 $\times$ 71.64 & 35.5k $\times$ 2 & 35.5k $\times$ 2 & 37.5k $\times$ 2 & 15.7k $\times$ 2 & 900k $\times$ 2 \\
Platform     & Drone, Ground, Satellite &  Ground, Satellite  &  Ground, Satellite & Ground, 45$^{\circ}$ Aerial & Ground, 45$^{\circ}$ Aerial & Ground, Satellite \\
\#imgs./location  &  54 + 16.64 + 1 &  1 + 1  & 1+1 & 1+1 &  1+1 &  1+1 \\
Target     & Building &  User &  User & Building & Building & User \\
GeoTag     & $\checkmark$  & $\checkmark$ &  $\checkmark$ &  $\checkmark$ & $\checkmark$ & $\checkmark$ \\
Evaluation  & Recall@K \& AP & Recall@K & Recall@K & PR curves \& AP & PR curves \& AP  & Recall@K \\
\hline
\end{tabular}
\end{center}
\caption{Comparison between University-1652 and other geo-localization datasets. The existing datasets usually consider matching the images from two platforms, and provide image pairs. In contrast, our dataset focuses on multi-view images, providing 71.64 images per location. 
For each benchmark, the table shows the number of training images and average images per location, as well as the availability of collection platform, geo-tag, and evaluation metric.
}
\vspace{-.2in}
\label{table:Dataset}
\end{table*}

\setlength{\tabcolsep}{12pt}
\begin{table}
\small
\begin{center}
\begin{tabular}{l|c|c|c}
\hline
Split & \#imgs & \#classes & \#universities\\
\shline
 Training & 50,218 & 701 & 33 \\
 \hline
 Query$_{drone}$ & 37,855 & 701 &  \multirow{6}{0.1\linewidth}{\centering{39}}\\
 Query$_{satellite}$& 701 & 701 & \\
 Query$_{ground}$ & 2,579 & 701 & \\
 Gallery$_{drone}$& 51,355 & 951 & \\
 Gallery$_{satellite}$ & 951 & 951 & \\
 Gallery$_{ground}$ & 2,921 & 793 & \\
\hline
\end{tabular}
\end{center}
\caption{Statistics of University-1652 training and test sets, including the image number and the building number of training set, query set and gallery set. We note that there is no overlap in the 33 universities of the training set and 39 universities of test sets.
}
\label{table:Statistics}
\end{table}

\section{University-1652 Dataset} \label{sec:dataset}
\subsection{Dataset Description}
In this paper, we collect  satellite-view images, drone-view images with the simulated drone cameras, and ground-view images for every location. We first select $1,652$ architectures of $72$ universities around the world as target locations. We do not select landmarks as the target. The two main concerns are: first, the landmarks usually contain discriminative architecture styles, which may introduce some unexpected biases; second, the drone is usually forbidden to fly around landmarks. Based on the two concerns, we select the buildings on the campus as the target, which is closer to the real-world practice. 

It is usually challenging to build the relation between images from different sources. Instead of collecting data and then finding the connections between various sources, we start by collecting the metadata. We first obtain the metadata of university buildings from Wikipedia \footnote{\url{https://en.wikipedia.org/wiki/Category:Buildings_and_structures_by_university_or_college}}, including building names and university affiliations. Second, we encode the building name to the accurate geo-location, \ie, latitude and longitude, by Google Map. We filter out the buildings with ambiguous search results, and there are $1,652$ buildings left. 
Thirdly, we project the geo-locations in Google Map to obtain the satellite-view images. For the drone-view images, due to the unaffordable cost of the real-world flight, we leverage the 3D models provided by Google Earth to simulate the real drone camera. The 3D model also provides manipulative viewpoints. To enable the scale changes and obtain comprehensive viewpoints, we set the flight curve as a spiral curve (see Figure~\ref{fig:spiral}(a)) and record the flight video with 30 frames per second. The camera flies around the target with three rounds. The height gradually decreases from $256$ meters to $121.5$ meters, which is close to the drone flight height in the real world \cite{rule2015airspace, brar2015drones}.

For ground-view images, we first collect the data from the street-view images near the target buildings from Google Map. Specifically, we manually collect the images in different aspects of the building (see Figure~\ref{fig:spiral}(b)). 
However, some buildings do not have the street-view photos due to the accessibility, \ie, most street-view images are collected from the camera on the top of the car. To tackle this issue, we secondly introduce one extra source, \ie, image search engine. We use the building name as keywords to retrieve the relevant images. However, one unexpected observation is that the retrieved images often contain lots of noise images, including indoor environments and duplicates. So we apply the ResNet-18 model trained on the Place dataset \cite{zhou2017places} to detect indoor images, and follow the setting in \cite{krause2016unreasonable} to remove the identical images that belong to two different buildings. 
In this way, we collect $5,580$ street-view images and $21,099$ common-view images from Google Map and Google Image, respectively.
It should be noted that images collected from Google Image only serve as an extra training set but a test set. 

Finally, every building has 1 satellite-view image, 1 drone-view video, and  $3.38$ real street-view images on average. We crop the images from the drone-view video every 15 frames, resulting in $54$ drone-view images. Overall, every building has totally $58.38$ reference images. Further, if we use the extra Google-retrieved data, we will have $16.64$ ground-view images per building for training. Compared with existing datasets (see Table \ref{table:Dataset}), we summarize the new features in University-1652 into the following aspects:

\noindent\textbf{1) Multi-source:} University-1652 contains the data from three different platforms, \ie, satellites, drones and phone cameras. To our knowledge, University-1652 is the first geo-localization dataset, containing drone-view images. 

\noindent\textbf{2) Multi-view:} University-1652 contains the data from different viewpoints. The ground-view images are collected from different facets of target buildings. Besides, synthetic drone-view images capture the target building from various distances and orientations. 

\noindent\textbf{3) More images per class:} Different from the existing datasets that provide image pairs, University-1652 contains $71.64$ images per location on average. During the training, more multi-source $\&$ multi-view data could help the model to understand the target structure as well as learn the viewpoint-invariant features. At the testing stage, more query images also enable the multiple-query setting. In the experiment, we show that multiple queries could lead to a more accurate target localization. 

\subsection{Evaluation Protocol}
The University-1652 has $1,652$ buildings in total. There are $1,402$ buildings containing all three views, \ie, satellite-view, drone-view and ground-view images, and $250$ buildings that lack either 3D model or street-view images. We evenly split the $1,402$ buildings into the training and test sets, containing 701 buildings of 33 Universities, 701 buildings of the rest 39 Universities. \textbf{We note that there are no overlapping universities in the training and test sets.} The rest 250 buildings are added to the gallery as distractors. More detailed statistics are shown in Table \ref{table:Statistics}. Several previous datasets \cite{liu2019lending,zhai2017predicting,vo2016localizing} adopt the Recall@K, whose value is $1$ if the first matched image has appeared before the $K$-th image. Recall@K is sensitive to the position of the first matched image, and suits for the test set with only one true-matched image in the gallery. In our dataset, however, there are multiple true-matched images of different viewpoints in the gallery. The Recall@K 
could not reflect the matching result of the rest ground-truth images. We, therefore, also adopt the average precision (AP) in \cite{lin2015learning,tian2017cross}. The average precision (AP) is the area under the PR (Precision-Recall) curve, considering all ground-truth images in the gallery. Besides Recall@K, we calculate the AP and report the mean AP value of all queries. 

\begin{figure}[t]
\begin{center}
     \includegraphics[width=1\linewidth]{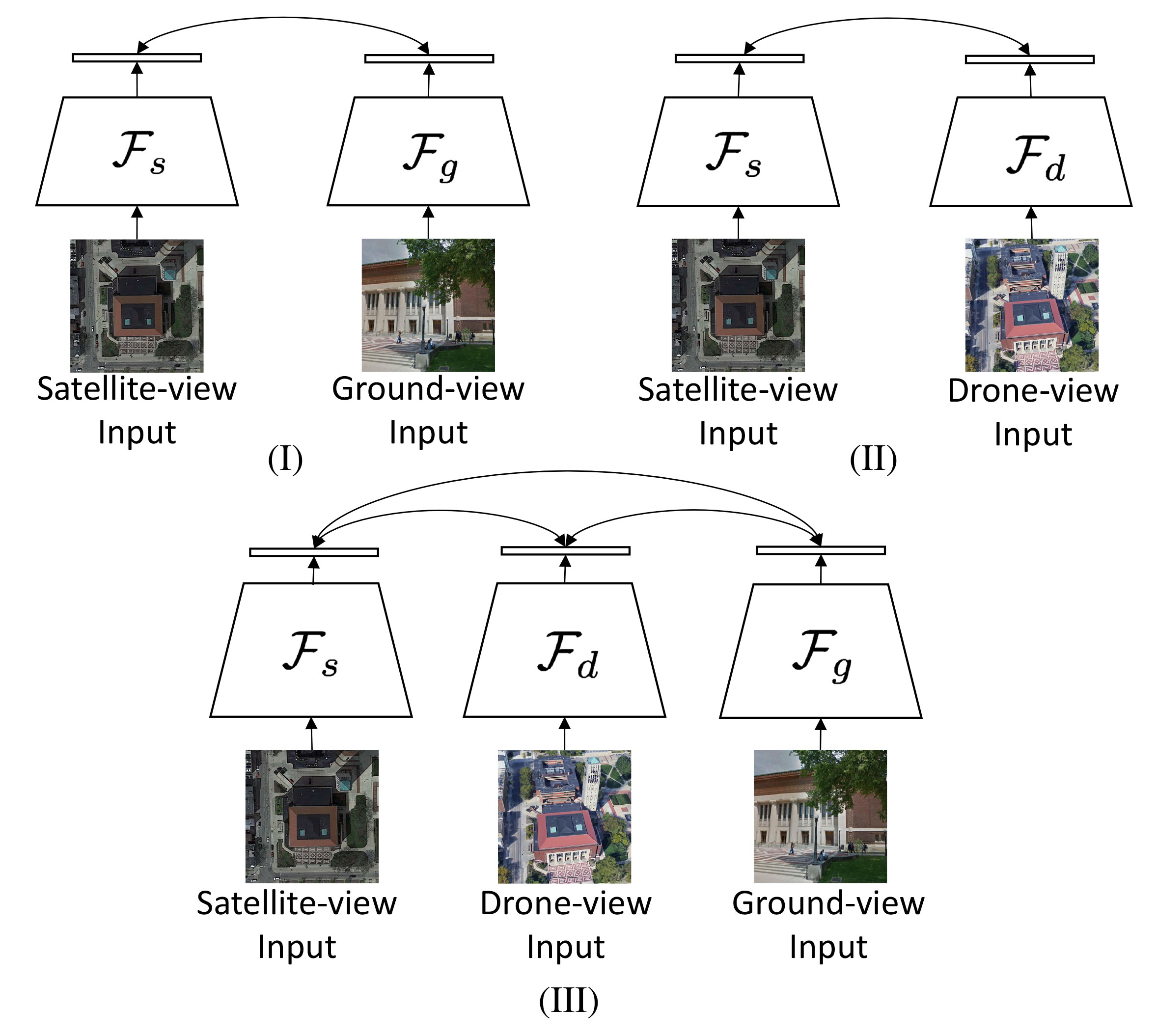}
\end{center}
\vspace{-.2in}
      \caption{ The basic model architectures for cross-view matching. Since the low-level patterns of different data are different, we apply multi-branch CNN to extract high-level features and then build the relation on the high-level features. (I) Model-I is a two-branch CNN model, which only considers the satellite-view and ground-view image matching; (II) Model-II is a two-branch CNN model, which only considers the satellite-view and drone-view image matching; (III) Model-III is a three-branch CNN model, which fully utilizes the annotated data, and considers the images of all three platforms. There are no ``standard'' methods to build the relationship between the data of multiple sources. Our baseline model applies the instance loss \cite{zheng2017dual} and we also could adopt other loss terms, \eg, triplet loss \cite{chechik2010large,deng2018triplet} and contrastive loss \cite{lin2015learning,workman2015wide,zheng2016discriminatively}.
      }\label{fig:method}
\end{figure}

\section{Cross-view Image Matching} \label{method}
Cross-view image matching could be formulated as a metric learning problem. The target is to map the images of different sources to a shared space. In this space, the embeddings of the same location should be close, while the embeddings of different locations should be apart. 

\subsection{Feature Representations}
There are no ``standard'' feature representations for the multi-source multi-view dataset, which demands robust features with good scalability towards different kinds of input images. In this work, we mainly compare two types of features: (1) the generic deep-learned features trained on extremely large datasets, such as ImageNet \cite{deng2009imagenet}, Place-365 \cite{zhou2017places}, and SfM-120k \cite{radenovic2018fine};  (2) the learned feature on our dataset. For a fair comparison, the backbone of all networks is ResNet-50 \cite{he2016deep} if not specified. More details are in Section~\ref{sec:localization}. Next, we describe the learning method on our data in the following section.

\subsection{Network Architecture and Loss Function}
The images from different sources may have different low-level patterns, so we denote three different functions $\mathcal{F}_s$, $\mathcal{F}_g$, and $\mathcal{F}_d$, which project the input images from satellites, ground cameras and drones to the high-level features. Specifically, to learn the projection functions, we follow the common practice in \cite{lin2015learning,liu2019lending}, and adopt the two-branch CNN as one of our basic structures. To verify the priority of the drone-view images to the ground-view images, we introduce two basic models for different inputs (see Figure~\ref{fig:method} (I),(II)). 
Since our dataset contains data from three different sources, we also extend the basic model to the three-branch CNN to fully leverage the annotated data (see Figure~\ref{fig:method} (III)). 

To learn the semantic relationship, we need one objective to bridge the gap between different views. 
Since our datasets provide multiple images for every target place, we could view every place as one class to train a classification model. In light of the recent development in image-language bi-directional retrieval, we adopt one classification loss called instance loss \cite{zheng2017dual} to train the baseline. The main idea is that a shared classifier could enforce the images of different sources mapping to one shared feature space. We denote $x_s$, $x_d$, and $x_g$ as three images of the location $c$, where $x_s$, $x_d$, and $x_g$ are the satellite-view image, drone-view image and ground-view image, respectively. 
Given the image pair $\{x_s, x_d\}$ from two views, the basic instance loss could be formulated as: 
\begin{align}
    p_s &= softmax(W_{share} \times \mathcal{F}_s(x_s)),\\
    L_s &= -\log(p_s(c)), \\
    p_d &= softmax(W_{share} \times \mathcal{F}_d(x_d)),\\
    L_d &= -\log(p_d(c)), 
\end{align}
where $W_{share}$ is the weight of the last classification layer. $p(c)$ is the predicted possibility of the right class $c$. Different from the conventional classification loss, the shared weight $W_{share}$ provides a soft constraint on the high-level features. We could view the $W_{share}$ as one linear classifier. 
After optimization, different feature spaces are aligned with the classification space.
In this paper, we further extend the basic instance loss to tackle the data from multiple sources. For example, if one more view is provided, we only need to include one more criterion term:
\begin{align}
    p_g &= softmax(W_{share} \times \mathcal{F}_g(x_g)), \\ 
    L_g &= -\log(p_g(c)), \\
    L_{total} &= L_s + L_d + L_g.
\end{align}
Note that we keep $W_{share}$ for the data from extra sources. In this way, the soft constraint also works on extra data. In the experiment, we show that the instance loss objective $L_{total}$ works effectively on the proposed University-1652 dataset. We also compare the instance loss with the widely-used triplet loss \cite{chechik2010large,deng2018triplet} and contrastive loss \cite{lin2015learning,workman2015wide,zheng2016discriminatively} with hard mining policy \cite{hermans2017defense,oh2016deep} in Section \ref{sec:ablation}.

\section{Experiment}
\subsection{Implementation Details}
We adopt the ResNet-50 \cite{he2016deep} pretrained on ImageNet \cite{deng2009imagenet} as our backbone model. We remove the original classifier for ImageNet and insert one $512$-dim fully-connected layer and one classification layer after the pooling layer. The model is trained by stochastic gradient descent with momentum $0.9$. The learning rate is $0.01$ for the new-added layers and $0.001$ for the rest layers. Dropout rate is $0.75$. While training, images are resized to $256 \times 256$ pixels. We perform simple data augmentation, such as horizontal flipping. For satellite-view images, we also conduct random rotation. When testing, we use the trained CNN to extract the corresponding features for different sources. The cosine distance is used to calculate the similarity between the query and candidate images in the gallery. The final retrieval result is based on the similarity ranking. 
If not specified, we deploy the Model-III, which fully utilizes the annotated data as the baseline model. We also share the weights of $\mathcal{F}_s$ and $\mathcal{F}_d$, since the two sources from aerial views share some similar patterns.

\setlength{\tabcolsep}{5pt}
\begin{table}
\small
\begin{center}
\begin{tabular}{l|c|cc|cc}
\hline
\multirow{2}{*}{Training Set} & Feature & \multicolumn{2}{c|}{Drone $\rightarrow$ Satellite} & \multicolumn{2}{c}{Satellite $\rightarrow$ Drone}\\
  & Dim & R@1 & AP & R@1 & AP\\
\shline
ImageNet \cite{deng2009imagenet} & 2048 & 10.11 & 13.04 & 33.24 & 11.59 \\
Place365 \cite{zhou2017places} & 2048 & 5.21 & 6.98 & 20.40 & 5.42 \\
SfM-120k \cite{radenovic2018fine} & 2048 & 12.53 & 16.08 & 37.09 & 10.28 \\
University-1652 & 512 & 58.49 & 63.13 & 71.18 & 58.74 \\
\hline
\end{tabular}
\end{center}
\caption{ Comparison between generic CNN features and the learned feature on the University-1652 dataset. The learned feature is shorter than the generic features but yields better accuracy. R@K (\%) is Recall@K, and AP (\%) is average precision (high is good).}
\label{table:Generic}
\end{table}

\setlength{\tabcolsep}{7pt}
\begin{table}
\small
\begin{center}
\begin{tabular}{l|cccc}
\hline
Query $\rightarrow$ Gallery & R@1 & R@5 & R@10 & AP\\
\shline
Ground $\rightarrow$ Satellite & 1.20 & 4.61 & 7.56 & 2.52 \\
Drone $\rightarrow$ Satellite & 58.49 & 78.67 & 85.23 & 63.13\\
\hline
$m$Ground $\rightarrow$ Satellite & 1.71 & 6.56 & 10.98 & 3.33 \\
$m$Drone $\rightarrow$ Satellite & 69.33 & 86.73 & 91.16 & 73.14 \\
\hline
\end{tabular}
\end{center}
\caption{Ground-view query vs. drone-view query. 
$m$ denotes multiple-query setting. The result suggests that drone-view images are superior to ground-view images when retrieving satellite-view images.}
\label{table:Street_vs_drone}
\end{table}

\subsection{Geo-localization Results} \label{sec:localization}
To evaluate multiple geo-localization settings, we provide query images from source $A$ and retrieve the relevant images in gallery $B$. We denote the test setting as $A \rightarrow B$.

\noindent\textbf{Generic features vs. learned features.} We evaluate two categories of features: 1) the generic CNN features. Some previous works \cite{workman2015wide} show that the CNN model trained on either ImageNet \cite{deng2009imagenet} or PlaceNet \cite{zhou2017places} has learned discriminative feature by itself. We extract the feature before the final classification layer. The feature dimension is $2048$. Besides, we also test the widely-used place recognition model \cite{radenovic2018fine}, whose backbone is ResNet-101. 
2) the CNN features learned on our dataset. Since we add one fully-connected layer before the classification layer, our final feature is $512$-dim. As shown in Table \ref{table:Generic}, our basic model achieves much better performance with the shorter feature length, which verifies the effectiveness of the proposed baseline. 

\noindent\textbf{Ground-view query vs. drone-view query.} We argue that drone-view images are more favorable comparing to ground-view images, since drone-view images are taken from a similar viewpoint, \ie, aerial view, with the satellite images. Meanwhile, drone-view images could avoid obstacles, \eg, trees, which is common in the ground-view images. To verify this assumption, we train the baseline model and extract the visual features of three kinds of data. As shown in Table \ref{table:Street_vs_drone}, when searching the relevant satellite-view images, the drone-view query is superior to the ground-view query. Our baseline model using drone-view query has achieved $58.49\%$ Rank@1 and $63.13\%$ AP accuracy. 

\noindent\textbf{Multiple queries.} Further, in the real-world scenario, one single image could not provide a comprehensive description of the target building. The user may use multiple photos of the target building from different viewpoints as the query. 
For instance, we could manipulate the drone fly around the target place to capture multiple photos. We evaluate the multiple-query setting by directly averaging the query features \cite{zheng2015scalable}. Searching with multiple drone-view queries generally arrives higher accuracy with about $10\%$ improvement in Rank@1 and AP, comparing with the single-query setting (see Table \ref{table:Street_vs_drone}). Besides, the target localization using the drone-view queries still achieves better performance than ground-view queries by a large margin. We speculate that the ground-view query does not work well in the single-query setting, which also limits the performance improvement in the multiple-query setting. 

\begin{figure}[t]
\begin{center}
     \includegraphics[width=1\linewidth]{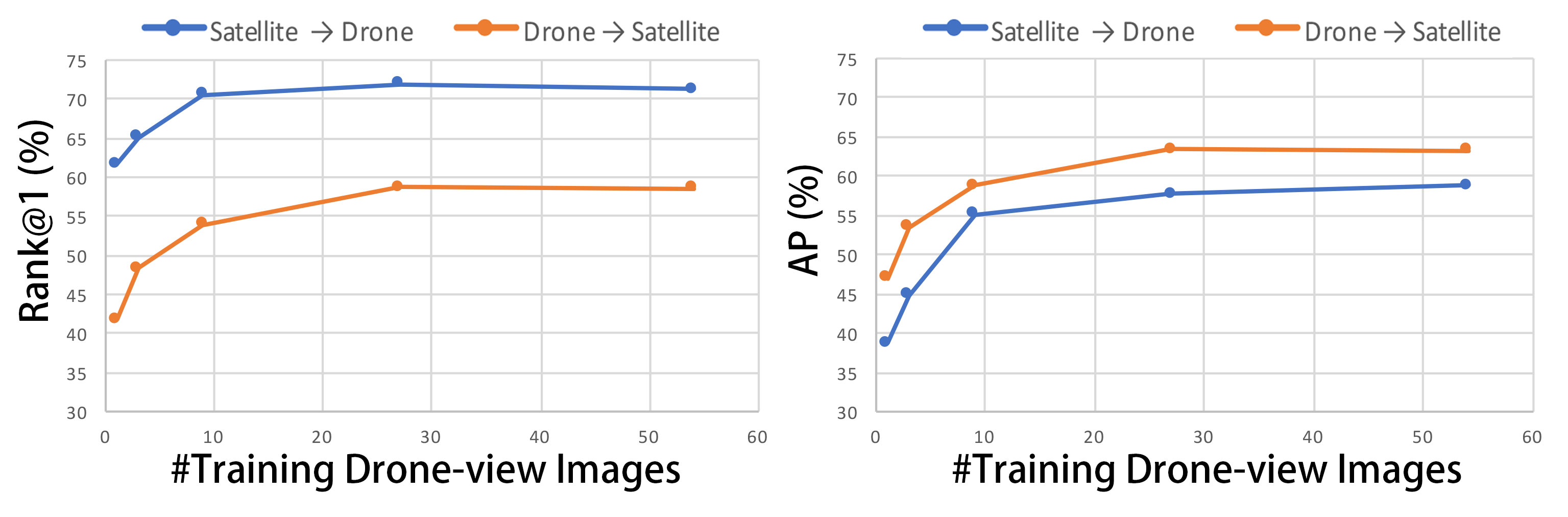}
\end{center}
\vspace{-.2in}
      \caption{ The test accuracy curves when using $n$ training drone-view images per class, $n\in\{1, 3, 9, 27, 54\}$. The two sub-figures are the Rank@1 (\%) and AP (\%) accuracy curves, respectively. The orange curves are for the drone navigation  (Satellite $\rightarrow$ Drone), and the blue curves are for the drone-view target localization (Drone $\rightarrow$ Satellite). 
      }\label{fig:views}
\end{figure}

\noindent\textbf{Does multi-view data help the viewpoint-invariant feature learning?} Yes. We fix the hyper-parameters and only modify the number of drone-view images in the training set. We train five models with $n$ drone-view images per class, where $n\in\{1,3,9,27,54\}$. As shown in Figure~\ref{fig:views}, when we gradually involve more drone-view training images from different viewpoints, the Rank@1 accuracy and AP accuracy both increase.  

\begin{figure}[t]
\begin{center}
     \includegraphics[width=1\linewidth]{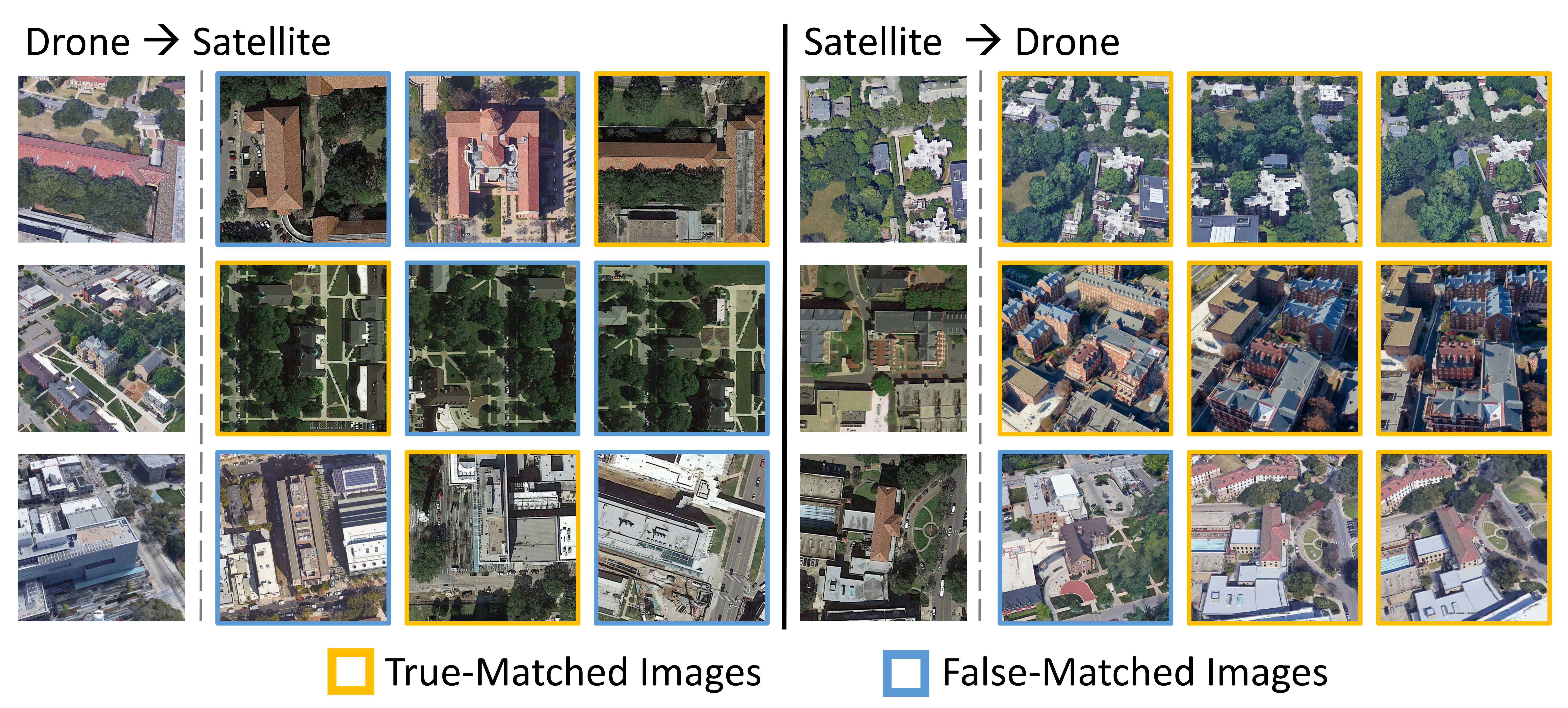}
\end{center}
\vspace{-.2in}
      \caption{ Qualitative image retrieval results. We show the top-3 retrieval results of drone-view target localization (left) and drone navigation (right). The results are sorted from left to right according to their confidence scores. The images in yellow boxes are the true matches, and the images in the blur boxes are the false matches. (Best viewed when zoomed in.)
      }\label{fig:visual}
\end{figure}

\begin{figure*}[t]
\begin{center}
     \includegraphics[width=1\linewidth]{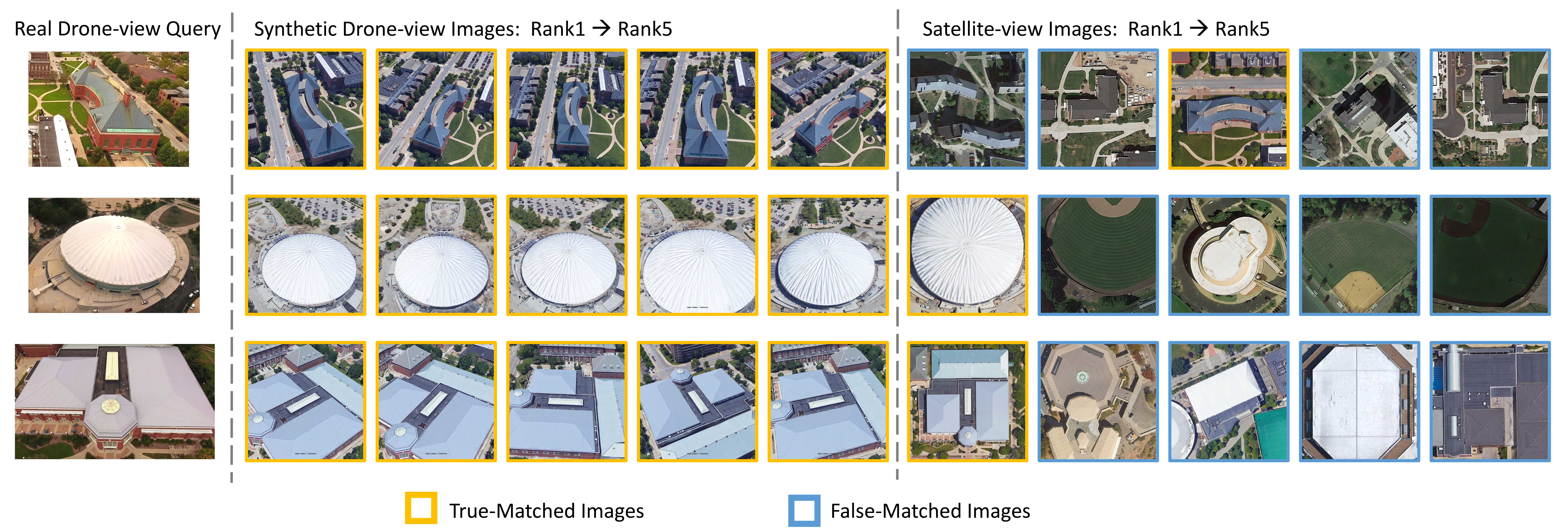}
\end{center}
\vspace{-.2in}
      \caption{ Qualitative image search results using real drone-view query. We evaluate the baseline model on an unseen university. There are two results: (I) In the middle column, we use the real drone-view query to search similar synthetic drone-view images. The result suggests that the synthetic data in University-1652 is close to the real drone-view images; (II) In the right column, we show the retrieval results on satellite-view images. It verifies that the baseline model trained on University-1652 has good generalization ability and works well on the real-world query.
      }\label{fig:real}
\end{figure*}

\noindent\textbf{Does the learned model work on the real data?} Yes. Due to the cost of collecting real drone-view videos, here we provide a qualitative experiment. We collect one 4K real drone-view video of University-X from Youtube granted by the author. University-X is one of the schools in the test set, and the baseline model has not seen any samples from University-X. We crop images from the video to evaluate the model. In Figure~\ref{fig:real}, we show the two retrieval results, \ie, Real Drone $\rightarrow$ Synthetic Drone, Real Drone $\rightarrow$ Satellite. The first retrieval result is to verify whether our synthetic data well simulates the images in the real drone cameras. We show the top-5 similar images in the test set retrieved by our baseline model. It demonstrates that the visual feature of the real drone-view query is close to the feature of our synthetic drone-view images.
The second result on the Real Drone $\rightarrow$ Satellite is to verify the generalization of our trained model on the real drone-view data. We observe that the baseline model has good generalization ability and also works on the real drone-view images for drone-view target localization. The true-matched satellite-view images are all retrieved in the top-5 of the ranking list.


\noindent\textbf{Visualization.} For additional qualitative evaluation, we show retrieval results by our baseline model on University-1652 test set (see Figure~\ref{fig:visual}). We can see that the baseline model is able to find the relevant images from different viewpoints. For the false-matched images, although they are mismatched, they share some similar structure pattern with the query image.

\setlength{\tabcolsep}{1pt}
\begin{table}
\small
\begin{center}
\begin{tabular}{l|cc|cc}
\hline
\multirow{2}{*}{Loss} & \multicolumn{2}{c|}{Drone $\rightarrow$ Satellite} & \multicolumn{2}{c}{Satellite $\rightarrow$ Drone}\\
  & R@1 & AP & R@1 & AP\\
\shline
Contrastive Loss & 52.39 & 57.44 & 63.91 & 52.24\\
Triplet Loss (margin$=$0.3)  & 55.18 & 59.97 & 63.62 & 53.85 \\
Triplet Loss (margin$=$0.5)  & 53.58 & 58.60 & 64.48 & 53.15 \\
Weighted Soft Margin Triplet Loss & 53.21 & 58.03 & 65.62 & 54.47\\
Instance Loss & 58.23 & 62.91 & 74.47 & 59.45 \\
\hline
\end{tabular}
\end{center}
\caption{Ablation study of different loss terms. To fairly compare the five loss terms, we trained the five models on satellite-view and drone-view data, and hold out the ground-view data. For contrastive loss, triplet loss and weighted soft margin triplet loss, we also apply the hard-negative sampling policy.
}
\label{table:loss}
\end{table}

\setlength{\tabcolsep}{9pt}
\begin{table}
\small
\begin{center}
\begin{tabular}{l|cc|cc}
\hline
\multirow{2}{*}{Method} & \multicolumn{2}{c|}{Drone $\rightarrow$ Satellite} & \multicolumn{2}{c}{Satellite $\rightarrow$ Drone}\\
  & R@1 & AP & R@1 & AP\\
\shline
Not sharing weights  & 39.84 & 45.91 & 50.36 & 40.71 \\
Sharing weights & 58.49 & 63.31 & 71.18 & 58.74 \\
\hline
\end{tabular}
\end{center}
\caption{Ablation study. With/without sharing CNN weights on University-1652. The result suggests that sharing weights could help to regularize the CNN model. 
}
\label{table:Share}
\end{table}

\setlength{\tabcolsep}{11pt}
\begin{table}
\small
\begin{center}
\begin{tabular}{l|cc|cc}
\hline
\multirow{2}{*}{Image Size}& \multicolumn{2}{c|}{Drone $\rightarrow$ Satellite} & \multicolumn{2}{c}{Satellite $\rightarrow$ Drone}\\
  & R@1 & AP & R@1 & AP\\
\shline
256 & 58.49 & 63.31 & 71.18 & 58.74 \\
384 & 62.99 & 67.69 & 75.75 & 62.09 \\
512 & 59.69 & 64.80 & 73.18 & 59.40 \\
\hline
\end{tabular}
\end{center}
\caption{Ablation study of different input sizes on the University-1652 dataset. 
}
\label{table:Size}
\end{table}

\setlength{\tabcolsep}{5pt}
\begin{table*}
\small
\begin{center}
\begin{tabular}{l|c|ccc|ccc|ccc|ccc}
\hline
\multirow{2}{*}{Model} & Training & \multicolumn{3}{c|}{Drone $\rightarrow$ Satellite} & \multicolumn{3}{c|}{Satellite $\rightarrow$ Drone} & \multicolumn{3}{c|}{Ground $\rightarrow$ Satellite} & 
\multicolumn{3}{c}{Satellite $\rightarrow$ Ground}\\
  & Set & R@1 & R@10 & AP & R@1 & R@10 & AP & R@1 & R@10 & AP & R@1 & R@10 & AP \\
\shline
Model-I & Satellite + Ground & - & - & - & - & - & - & 0.62 & 5.51 & 1.60 & 0.86 & 5.99 & 1.00 \\
Model-II & Satellite + Drone & 58.23 & 84.52 & 62.91 & 74.47 & 83.88 & 59.45 & - & - & - & - & - & -\\
Model-III & Satellite + Drone + Ground & 58.49 & 85.23 &  63.13 & 71.18 & 82.31 & 58.74 & 1.20 & 7.56 & 2.52 & 1.14 & 8.56 & 1.41 \\
\hline
\end{tabular}
\end{center}
\caption{Comparison of the three CNN models mentioned in Figure~\ref{fig:method}. R@K (\%) is Recall@K, and AP (\%) is average precision (high is good). Model-III that utilizes all annotated data outperforms the other two models in the three of four tasks.
}
\vspace{-.2in}
\label{table:Baseline}
\end{table*}

\subsection{Ablation Study and Further Discussion} \label{sec:ablation}
\noindent\textbf{Effect of loss objectives.}
The triplet loss and contrastive loss are widely applied in previous works \cite{lin2015learning,workman2015wide,deng2018triplet,chechik2010large,zheng2016discriminatively}, and the weighted soft margin triplet loss is deployed in \cite{hu2018cvm,liu2019lending,cai2019ground}. We evaluate these three losses on two tasks, \ie, Drone $\rightarrow$ Satellite and Satellite $\rightarrow$ Drone and compare three losses with the instance loss used in our baseline. For a fair comparison, all losses are trained with the same backbone model and only use drone-view and satellite-view data as the training set. For the triplet loss, we also try two common margin values $\{0.3, 0.5\}$. 
In addition, the hard sampling policy is also applied to these baseline methods during training \cite{hermans2017defense,oh2016deep}. As shown in Table \ref{table:loss}, we observe that the model with instance loss arrives better performance than the triplet loss and contrastive loss on both tasks.

\noindent\textbf{Effect of sharing weights.} In our baseline model, $\mathcal{F}_s$ and $\mathcal{F}_d$ share weights, since two aerial sources have some similar patterns. We also test the model without sharing weights (see Table \ref{table:Share}). The performance of both tasks drops. The main reason is that limited satellite-view images (one satellite-view image per location) are prone to be overfitted by the separate CNN branch. When sharing weights, drone-view images could help regularize the model, and the model, therefore, achieves better Rank@1 and AP accuracy.

\noindent\textbf{Effect of the image size.} Satellite-view images contain the fine-grained information, which may be compressed with small training size. We, therefore, try to enlarge the input image size and train the model with the global average pooling. The dimension of the final feature is still $512$. As shown in Table \ref{table:Size}, when we increase the input size to $384$, the accuracy of both task, drone-view target localization (Drone $\rightarrow$ Satellite) and drone navigation (Satellite $\rightarrow$ Drone) increases. However, when we increase the size to $512$, the performance drops. We speculate that the larger input size is too different from the size of the pretrained weight on ImageNet, which is $224\times224$. As a result, the input size of $512$ does not perform well. 

\noindent\textbf{Different baseline models.} 
We evaluate three different baseline models as discussed in Section~\ref{method}. As shown in Table \ref{table:Baseline}, there are two main observations: 
1). Model-II has achieved better Rank@1 and AP accuracy for drone navigation (Satellite $\rightarrow$ Drone). It is not surprising since Model-II is only trained on the drone-view and satellite-view data. 2). Model-III, which fully utilizes all annotated data, has achieved the best performance in the three of all four tasks. It could serve as a strong baseline for multiple tasks. 

\noindent\textbf{Proposed baseline on the other benchmark.} 
As shown in Table~\ref{table:usa}, we also evaluate the proposed baseline on one widely-used two-view benchmark, \eg, CVUSA~\cite{zhai2017predicting}. For fair comparison, we also adopt the 16-layer VGG~\cite{simonyan2014very} as the backbone model. We do not intend to push the state-of-the-art performance but to show the flexibility of the proposed baseline, which could also work on the conventional dataset. We, therefore, do not conduct tricks, such as image alignment~\cite{shi2019spatial} or feature ensemble~\cite{regmi2019bridging}.
Our intuition is to provide one simple and flexible baseline to the community for further evaluation. 
Compared with the conventional Siamese network with triplet loss, the proposed method could be easily extended to the training data from $N$ different sources ($N\geq2$). 
The users only need to modify the number of CNN branches. 
Albeit simple, the experiment verifies that the proposed method could serve as a strong baseline and has good scalability toward real-world samples.

\setlength{\tabcolsep}{9pt}
\begin{table}
\small
\begin{center}
\begin{tabular}{l|cccc}
\hline
Methods  & R@1 & R@5 & R@10 & R@Top1\%\\
\shline
Workman \cite{workman2015wide} & - & - & - & 34.40 \\
Zhai \cite{zhai2017predicting} & - & - & - & 43.20 \\
Vo \cite{vo2016localizing} & - & - & - & 63.70 \\
CVM-Net \cite{hu2018cvm} & 18.80 & 44.42 & 57.47 & 91.54 \\
Orientation \cite{liu2019lending}$^\dagger$ & 27.15 & 54.66 & 67.54 & \textbf{93.91} \\
Ours  & \textbf{43.91} & \textbf{66.38} & \textbf{74.58} & 91.78 \\
\hline
\end{tabular}
\end{center}
\caption{ Comparison of results on the two-view dataset CVUSA~\cite{zhai2017predicting} with VGG-16 backbone. $^\dagger$: The method utilizes extra orientation information as input.
}
\label{table:usa}
\end{table}

\setlength{\tabcolsep}{3pt}
\begin{table}
\small
\begin{center}
\begin{tabular}{l|c|c|c|c|c|c}
\hline
Method & Oxford	& Paris	& ROxf (M) & RPar (M) & ROxf (H) & RPar (H) \\
\shline
ImageNet & 3.30 & 6.77 & 4.17 & 8.20 & 2.09 & 4.24 \\
$\mathcal{F}_s$ & 9.24 & 13.74 & 5.83 & 13.79 & 2.08 & 6.40\\
$\mathcal{F}_g$ & 25.80 & 28.77 & 15.52 & 24.24 & 3.69 & 10.29\\
\hline
\end{tabular}
\end{center}
\caption{Transfer learning from University-1652 to small-scale datasets. We show the AP (\%) accuracy on Oxford \cite{philbin2007object}, Paris \cite{philbin2008lost}, ROxford and RParis \cite{RITAC18}. For ROxford and RParis, we report results in both medium (M) and hard (H) settings. 
}
\label{table:transfer}
\end{table}

\noindent\textbf{Transfer learning from University-1652 to small-scale datasets.} 
We evaluate the generalization ability of the baseline model on two small-scale datasets, \ie, Oxford \cite{philbin2007object} and Pairs \cite{philbin2008lost}. Oxford and Pairs are two popular place recognition datasets. We directly evaluate our model on these two datasets without finetuning. Further, we also report results on the revised Oxford and Paris datasets (denoted as ROxf and RPar) \cite{RITAC18}. In contrast to the generic feature trained on ImageNet \cite{deng2009imagenet}, the learned feature on University-1652 shows better generalization ability. Specifically, we try two different branches, \ie, $\mathcal{F}_s$ and $\mathcal{F}_g$, to extract features. $\mathcal{F}_s$ and $\mathcal{F}_g$ share the high-level feature space but pay attention to different low-level patterns of inputs from different platforms. $\mathcal{F}_s$ is learned on satellite-view images and drone-view images, while $\mathcal{F}_g$ learns from ground-view images. As shown in Table \ref{table:transfer}, $\mathcal{F}_g$ has achieved better performance than $\mathcal{F}_s$. We speculate that there are two main reasons. First, the test data in Oxford and Pairs are collected from Flickr, which is closer to the Google Street View images and the images retrieved from Google Image in the ground-view data. Second, $\mathcal{F}_s$ pay more attention to vertical viewpoint changes instead of horizontal viewpoint changes, which are common in Oxford and Paris.

\section{Conclusion}
This paper contributes a multi-view multi-source benchmark called University-1652. University-1652 contains the data from three platforms, including satellites, drones and ground cameras, and enables the two new tasks, \ie, drone-view target localization and drone navigation. We view the two tasks as the image retrieval problem, and present the baseline model to learn the viewpoint-invariant feature. In the experiment, we observe that the learned baseline model has achieved competitive performance towards the generic feature, and shows the feasibility of drone-view target localization and drone navigation. In the future, we will continue to investigate more effective and efficient feature of the two tasks. 

\noindent\textbf{Acknowledgement.} We would like to thank the real drone-view video providers, Regal Animus and FlyLow. Their data helps us verify our dataset and models. 

{
\bibliographystyle{ACM-Reference-Format}
\bibliography{egbib}
}

\clearpage
\appendix
\section*{Appendix}
\setcounter{section}{0}
\renewcommand\thesection{\Alph{section}}

\section{More Quantitative Results}


\textbf{With/without Google Image data.} In the University-1652 training data, we introduce the ground-view images collected from Google Image. We observe that although the extra data retrieved from Google Image contains noise, most images are true-matched images of the target building. In Table \ref{table:Google}, we report the results with/without the Google Image training data. The baseline model trained with extra data generally boosts the performance not only on the ground-related tasks, \ie, Ground $\rightarrow$ Satellite and Satellite $\rightarrow$ Ground,  but on the drone-related tasks, \ie, Drone $\rightarrow$ Satellite and Satellite $\rightarrow$ Drone. The result also verifies that our baseline method could perform well against the noisy data in the training set.

\section{More Qualitative Results}
\begin{figure*}[t]
\begin{center}
    \includegraphics[width=0.95\linewidth]{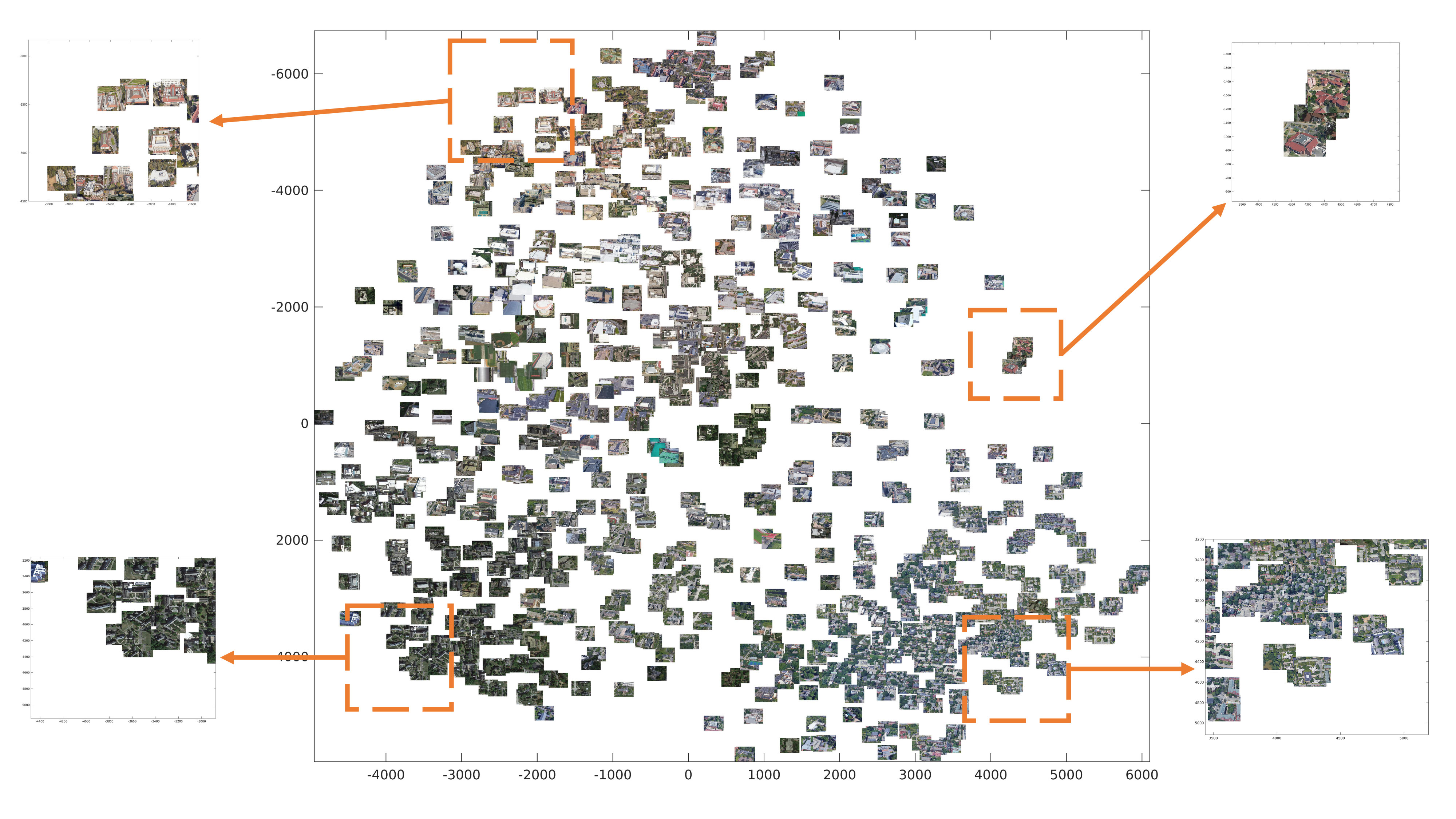}
\end{center}
     \caption{Visualization of cross-view features using t-SNE \cite{van2014accelerating} on University-1652. (Best viewed when zoomed in.)
     }\label{fig:cluster}
\end{figure*}

\textbf{Visualization of cross-view features.} 
We sample $500$ pairs of drone-view and satellite-view images in the test set to extract features and then apply the widely-used t-SNE \cite{van2014accelerating} to learn the 2D projection of every feature. As shown in Figure~\ref{fig:cluster}, the features of the same location are close, and the features of different target buildings are far away. It demonstrates that our baseline model learns the effective feature space, which is discriminative.

\section{More Details of University-1652}

\textbf{Building name list.} We show the name of the first $100$ buildings in University-1652 (see Table \ref{table:BuildingName}). 

\noindent\textbf{Data License.} We carefully check the data license from Google. There are two main points.
First, the data of Google Map and Google Earth could be used based on fair usage. We will follow the guideline on this official website \footnote{\url{https://www.google.com/permissions/geoguidelines/}}.
Second, several existing datasets have utilized the Google data. In practice, we will adopt a similar policy of existing datasets \footnote{\url{http://www.ok.ctrl.titech.ac.jp/~torii/project/247/}}$^,$ 
\footnote{\url{http://cs.uky.edu/~jacobs/datasets/cvusa/}} to release the dataset based on the academic request. 

\begin{figure}[t]
\begin{center}
    \includegraphics[width=1\linewidth]{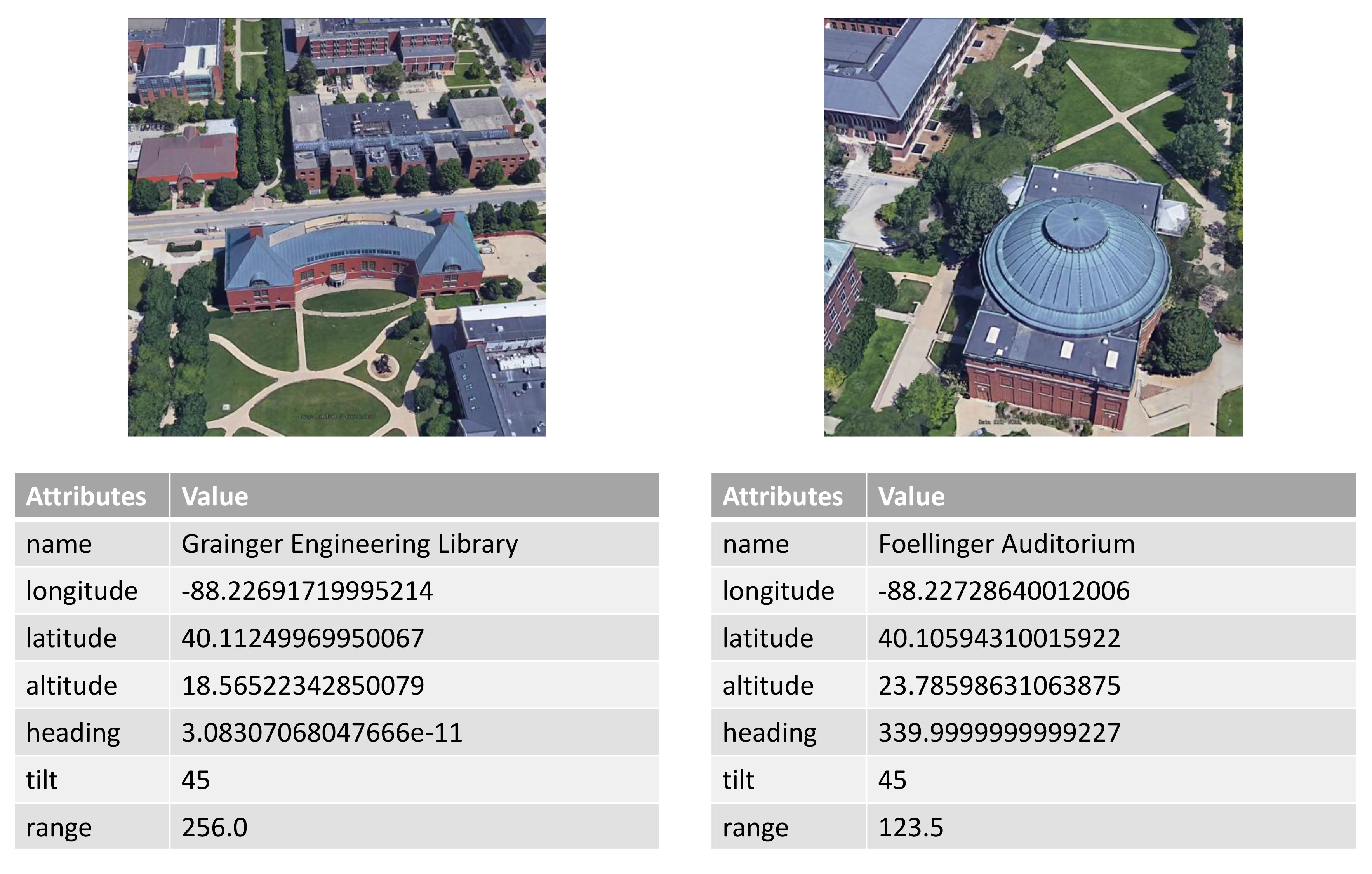}
\end{center}
     \caption{Metadata samples. We record attributes for every frame, including the building name, longitude, latitude, altitude, heading, tilt and range. }\label{fig:metadata}
\end{figure}

\noindent\textbf{Frame-level metadata.} Besides drone-view videos, we also record the frame-level metadata, including the building name, longitude, latitude, altitude, heading, tilt and range (see Figure~\ref{fig:metadata}). Exploiting metadata is out of the scope of this paper, so we do not explore the usage of attributes in this work. But we think that the metadata could enable the future study, \eg, orientation alignment between drone-view images and satellite-view images. In the future, we will continue to study this problem.



\setlength{\tabcolsep}{7pt}
\begin{table*}
\small
\begin{center}
\begin{tabular}{l|ccc|ccc|ccc|ccc}
\hline
\multirow{2}{*}{Model} & \multicolumn{3}{c|}{Drone $\rightarrow$ Satellite} & \multicolumn{3}{c|}{Satellite $\rightarrow$ Drone} & \multicolumn{3}{c|}{Ground $\rightarrow$ Satellite} & 
\multicolumn{3}{c}{Satellite $\rightarrow$ Ground}\\
  & R@1 & R@10 & AP & R@1 & R@10 & AP & R@1 & R@10 & AP & R@1 & R@10 & AP \\
\shline
Without noisy data  & 57.52 & 83.89 & 62.29 & 69.19 & 82.31 & 56.15 & \textbf{1.28} & 6.20 & 2.29& \textbf{1.57} & 7.13 & \textbf{1.52}\\
With noisy data & \textbf{58.49} & \textbf{85.23} &  \textbf{63.13} & \textbf{71.18} & \textbf{82.31} & \textbf{58.74} & 1.20 & \textbf{7.56} & \textbf{2.52} & 1.14 & \textbf{8.56} & 1.41 \\
\hline
\end{tabular}
\end{center}
\caption{Ablation study. With / without noisy training data from Google Image. The baseline model trained with the Google Image data is generally better in all four tasks. The result also verifies that our baseline method could perform well against the noise in the dataset.
}
\label{table:Google}
\end{table*}

  
\setlength{\tabcolsep}{1pt}
\begin{table*}
\small
\begin{center}
\begin{tabular}{c|c}
\hline
\multicolumn{2}{c}{Building Names}\\
\shline
Bibliothèque Saint-Jean, University of Alberta & Clare Drake Arena\\
Foote Field & Myer Horowitz Theatre\\
National Institute for Nanotechnology & St Joseph's College, Edmonton\\
Stollery Children's Hospital & Universiade Pavilion, University of Alberta\\
University of Alberta Hospital & Alberta B. Farrington Softball Stadium\\
Decision Theater, University of Alberta & Gammage Memorial Auditorium\\
Harrington–Birchett House & Industrial Arts Building\\
Irish Field & Louise Lincoln Kerr House and Studio\\
Matthews Hall, University of Alberta & Mona Plummer Aquatic Center\\
Old Main (Arizona State University) & Packard Stadium, University of Alberta\\
Security Building (Phoenix, Arizona) & Sun Devil Gym, University of Alberta\\
Sun Devil Stadium, University of Alberta & United States Post Office (Phoenix, Arizona)\\
Wells Fargo Arena (Tempe, Arizona) & Administration Building, University of Alberta\\
Wheeler Hall, University of Alberta & Marting Hall, University of Alberta\\
Malicky Center, University of Alberta & Burrell Memorial Observatory\\
Kleist Center for Art and Drama & Wilker Hall, University of Alberta\\
Kamm Hall, University of Alberta & Dietsch Hall, University of Alberta\\
Telfer Hall, University of Alberta & Ward Hall, University of Alberta\\
Thomas Center for Innovation and Growth (CIG) & Kulas Musical Arts Building, Baldwin Wallace University\\
Boesel Musical Arts Center, Baldwin Wallace University & Merner-Pfeiffer Hall, Baldwin Wallace University\\
Ritter Library, Baldwin Wallace University & Lindsay-Crossman Chapel, Baldwin Wallace University\\
Presidents House, Baldwin Wallace University & Student Activities Center (SAC), Baldwin Wallace University\\
Strosacker Hall (Union), Baldwin Wallace University & Bonds Hall, Baldwin Wallace University\\
Durst Welcome Center, Baldwin Wallace University & Lou Higgins Center, Baldwin Wallace University\\
Tressel Field @ Finnie Stadium, Baldwin Wallace University & Rutherford Library\\
Rudolph Ursprung Gymnasium, Baldwin Wallace University & Packard Athletic Center (formerly Bagley Hall), Baldwin Wallace University\\
Baldwin-Wallace College North Campus Historic District & Baldwin-Wallace College South Campus Historic District\\
Binghamton University Events Center, Binghamton University & Commonwealth Avenue, Boston University\\
Boston University Photonics Center, Boston University & Boston University School of Law, Boston University\\
Boston University Track and Tennis Center, Boston University & Boston University West Campus\\
BU Castle, Boston University & George Sherman Union, Boston University\\
John Hancock Student Village, Boston University & Marsh Chapel, Boston University\\
Metcalf Center for Science and Engineering, Boston University & Morse Auditorium, Boston University\\
Mugar Memorial Library, Boston University & Myles Standish Hall, Boston University\\
Questrom School of Business, Boston University & Shelton Hall (Boston University), Boston University\\
Walter Brown Arena, Boston University & Warren Towers, Boston University\\
Benson (Ezra Taft) Building, Brigham Young University & Brimhall (George H.) Building, Brigham Young University\\
BYU Conference Center, Brigham Young University & Centennial Carillon Tower, Brigham Young University\\
Chemicals Management Building, Brigham Young University & Clark (Herald R.) Building, Brigham Young University\\
Clark (J. Reuben) Building (Law School), Brigham Young University & Clyde (W.W.) Engineering Building, Brigham Young University\\
Crabtree (Roland A.) Technology Building, Brigham Young University & Eyring (Carl F.) Science Center, Brigham Young University\\
Faculty Office Building, Brigham Young University & Former Presidents' Home, Brigham Young University\\
BYU Testing Center, Grant (Heber J.) Building, Brigham Young University & Harman (Caroline Hemenway) Building, Brigham Young University\\
Harris (Franklin S.) Fine Arts Center, Brigham Young University & Kimball (Spencer W.) Tower, Brigham Young University\\
Knight (Amanda) Hall, Brigham Young University & Knight (Jesse) Building, Brigham Young University\\
Lee (Harold B.) Library, Brigham Young University & Life Sciences Building, Brigham Young University\\
Maeser (Karl G.) Building, Brigham Young University & Martin (Thomas L.) Building, Brigham Young University\\
McKay (David O.) Building, Brigham Young University & Smith (Joseph F.) Building, Brigham Young University\\
Smith (Joseph) Building, Brigham Young University & Snell (William H.) Building, Brigham Young University\\
Talmage (James E.) Math Sciences/Computer Building, Brigham Young University & Tanner (N. Eldon) Building, Brigham Young University\\
\hline
\end{tabular}
\end{center}
\caption{Due to the space limitation, here we show the first $100$ building names in the University-1652 dataset.
}
\label{table:BuildingName}
\end{table*}

\clearpage

\end{document}